# Measuring the Recyclability of Electronic Components to Assist Automatic Disassembly and Sorting Waste Printed Circuit Boards

Muhammad Mohsin[1, *], Xianlai Zeng[2], Stefano Rovetta[1,3], Francesco Masulli[1,3]

(1. DIBRIS University of Genoa, Via Dodecaneso 35, Genoa, 16146, Italy 2. School of Environment, Tsinghua University, Beijing 100084, China 3. Vega Research Laboratories s.r.l. , Genoa, 16121, Italy)
*Email: muhammad.mohsin@edu.unige.it

**Abstract**
The waste of electrical and electronic equipment has been increased due to the fast evolution of technology products and competition of many IT sectors. Every year millions of tons of electronic waste are thrown into the environment which causes high consequences for human health. Therefore, it is crucial to control this waste flow using technology, especially using Artificial Intelligence but also reclamation of critical raw materials for new production processes. In this paper, we focused on the measurement of recyclability of waste electronic components (WECs) from waste printed circuit boards (WPCBs) using mathematical innovation model. This innovative approach evaluates both the recyclability and recycling difficulties of WECs, integrating an AI model for improved disassembly and sorting. Assessing the recyclability of individual electronic components present on WPCBs provides insight into the recovery potential of valuable materials and indicates the level of complexity involved in recycling in terms of economic worth and production utility. This novel measurement approach helps AI models in accurately determining the number of classes to be identified and sorted during the automated disassembly of discarded PCBs. It also facilitates the model in iterative training and validation of individual electronic components.

## 1    Introduction

The issue of waste of electrical and electronic equipment (WEEE or e-waste) is indeed a significant concern due to its rapid increase worldwide. This waste includes a wide range of devices such as mobile phones, computers, televisions, fridges, household appliances, lamps, medical devices, and photovoltaic panels. E-waste contains hazardous materials that can cause major environmental and health problems if not managed properly. Moreover, modern electronics contain rare and expensive resources that can be recycled and reused, contributing to a circular economy, and supporting the security of supply for critical raw materials [1]. The United Nations reported that in 2019, the world produced 53.6 million tons of e-waste, which was worth $62.5 billion. However, only 17% of this e-waste was properly recycled, while the rest ended up in landfills or was recycled improperly. This situation is concerning due to the significant environmental and health risks associated with improper e-waste management. The report also predicted that with the continuous increasing trend, the annual global e-waste generation would reach 120 million tons by 2050 [2,3]. E-waste is highly complicated since it contains up to 1000 chemicals and 60 elements, such as plastics, epoxy, ceramics, glass, and different types of metals, including base metals, precious metals, and rare earth metals [4–6]. The recycling technologies for these materials exhibit significant diversity, demanding the need to narrow the attention to a specific category. Therefore, in this research article, we focused on one category of e-waste called waste printed circuit boards (WPCBs).

E-waste consists mostly of WPCBs containing valuable materials. These PCBs are sourced from obsolete mobile phones, smartphones, and computers [7–9]. WPCBs are crucial in the recycling of WEEE due to their long-lasting usefulness. WPCBs contain various metals, including copper, silver, gold, and palladium, as well as base metals like Cu, Fe, Ni, Zn, Sn, and Pb. These metals pose economic and environmental hazards if not properly recycled. Recycling PCBs helps reduce environmental impact and conserve resources [10–12]. Several technologies have been established to recycle and retrieve valuable materials from electronic waste. These include traditional methods like pyrometallurgy, hydrometallurgy, and biohydrometallurgy, as well as emerging methods like supercritical fluid extraction, flash joule heating, and photocatalysis [13–16]. These traditional and emerging technologies used for recycling of e-waste, mainly focused on bare boards PCBs due to their ease of recycling as well as high concentrations of metals present in them. Previously, the recycling of WPCBs has mostly targeted bare boards since they are easier to recycle and contain higher amounts of metals [11][17,18]. However, the electronic components of WPCBs, such as integrated circuits, electrolytic capacitors, ceramic capacitors, transistors, coils, resistors, and other electronic parts, have been largely neglected in conventional recycling methods. The reason for not focusing on electronic components is due to the complexity and high cost associated with their disassembly, sorting and recycling processes. Further, manual disassembly is not effective at large scale industrial level recycling. Electronic components often include valuable critical raw materials including rare-earth elements. However, extracting these materials is challenging and requires specialized equipment and expertise. This led to a potential gap which needs special attention in terms of intelligent recycling solutions.

The first step in the process of electronic component disassembly and sorting from WPCBs using Artificial Intelligent based systems is to measure and assess the recyclability of each individual component based on their material composition as well as economic valuation. AI based systems require a huge amount of data for their smooth implementation which takes lot of time and resources during prototyping. Fig. 1 shows the example images of WPCBs high resolution images. Therefore, in this research, we measure the recyclability of individual components present on the WPCBs using an innovative model based on different factors. This study will not only help the AI systems but also speed up the close loop system of intelligent disassembly by characterization of individual components as class during training and inference.

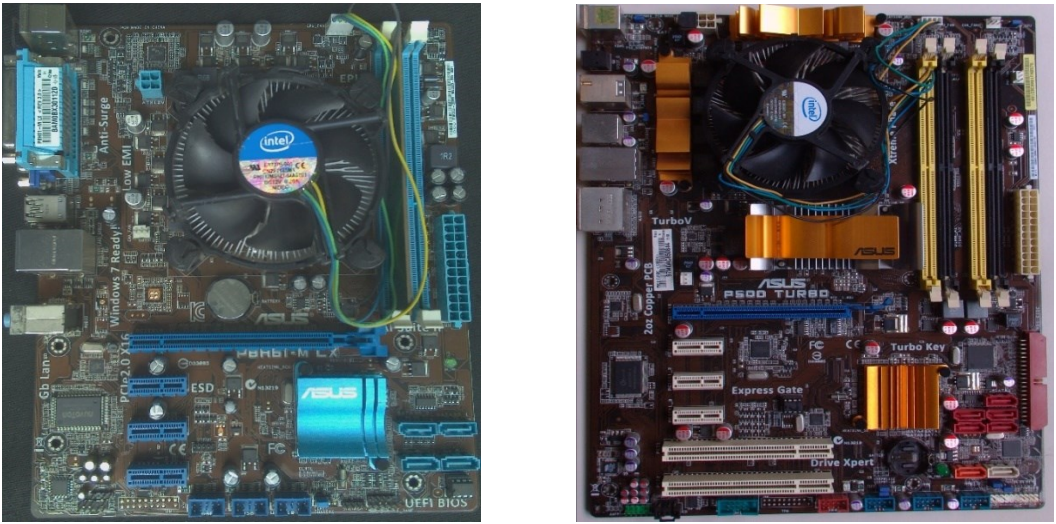

Fig. 1. Examples of high resolution WPCBs images from V-PCBs dataset

## 2 Methods

*2.1 Current Status of Waste Printed Circuit Board*

PCBs are usually made-up resins with fiber glass and lamination of copper foil for the connectivity of different electronic components on the boards. There are different types of PCBs, for instance single side, double sided and multilayered PCBs. When a device's efficiency rises, the number of layers increases, as in the case of double- and multilayered PCBs. The composition in overall PCBs varies depending on the kind of PCBs (electric or electronic), year of manufacture [19,20]. For instance, the evaluation of different metals and their composition is investigated [21] and its results clarify that the concentrations of metals in printed circuit boards (PCBs) of cathode ray tube televisions (TV-PCBs) produced between 1980 and 2005, with a focus on Pb, As, Cu, Au, Sn, and Ag. The results show that there were no major modifications in the compositions of Pb and Sn over the years, suggesting no significant changes in the Sn/Pb solder used in joining the circuitry system. The average composition of Cu fluctuated between 5.10% in 1980 and 12.8% in the mid-1990s, decreasing afterwards, possibly due to thinner layers of Cu in newer model products. Due to this variation, more innovative methods are crucial to efficiently recycle the WPCBs.

Waste PCBs comprise various electronic components that can be classified according to their function and material composition. The electronic components have different sizes and shapes which are mounted on PCBs surface. For instance, electrolytic capacitor, integrated circuit, resistor, and transistor. Research on recycling WPCBs has primarily concentrated on extracting valuable metals such as copper, gold, and other precious metals from the bare boards [22,23]. However, there has been little attention given to reusing the electronic components connected to the PCBs, even though electronic components on WPCBs generally retain a considerable amount of usable life and precious materials that can be utilized in manufacturing processes [24].

The reusability and efficient recycling of WECs relies on the disassembly process, which can be manual, semi-automated, or fully automatic. Current recycling processes often directly crush WPCBs along with components to recover major metals, which does not represent the real situation and potential for component reuse. Manual disassembly is labor-intensive and costly [25]. Developing environmentally safe and high-efficiency strategies for removing electronic components (EECs) from WPCBs is crucial for their crushing and metal recovery [26]. Many studies have been conducted on automating the dismantling and sorting of WECs to minimize environmental impact and enhance feasibility [27].

*2.2 Automatic Disassembly of WECs and sorting using Artificial Intelligence.*

Artificial Intelligence based models have the potential to recognize the different WECs on PCBs boards and sort them in similar categories using robotic manipulator. For best fostering the AI potential, models require huge amount of data and computational resources for their training, validation, and inference in real time scenarios [28].

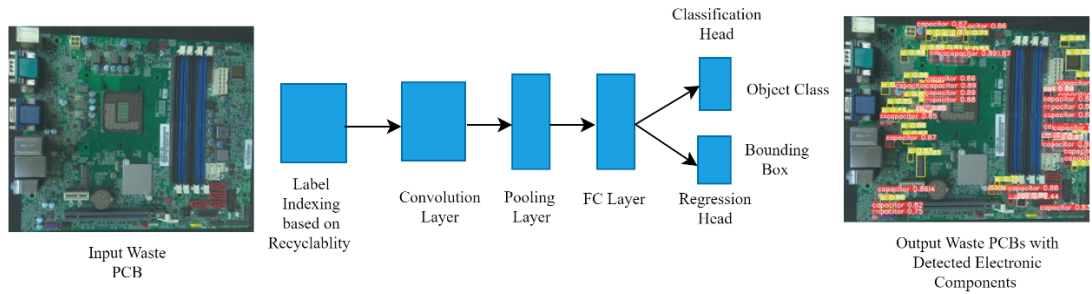

**Fig. 2. Object Detection Model for WECs from WPCBs based on high recyclability.**

Fig. 2 illustrates that the initial stage in a deep learning model is to input images, where in the case of WPCBs, the input images are WPCBs with their respective labels or classes. For instance, capacitor, resistor, and IC. Choosing the object class is crucial because the model requires a significant amount of energy and computational resources for both training and validation. For targeted recycling, we use AI to focus on electronic components. These high-end components are utilized as input classes for the convolution process, which is then followed by pooling and fully connected layers for classifying the position of the class in the form of a bounding box.

## 3　Recyclability Assessment: An Innovative Methodological Approach

### 3.1 Recyclability definition and its core components in general

The term recyclability in e-waste refers to measuring the extent of materials and components used with in waste printed circuit boards (WPCBs) can be recycled and re-used in new production processes. Measuring recyclability involves assessing the ease and efficiency with which the PCBs can be disassembled, separated into valuable components, and processed for reuse or material recovery. Fig. 3 shows the key components involved in measuring the overall recyclability of electronic waste.

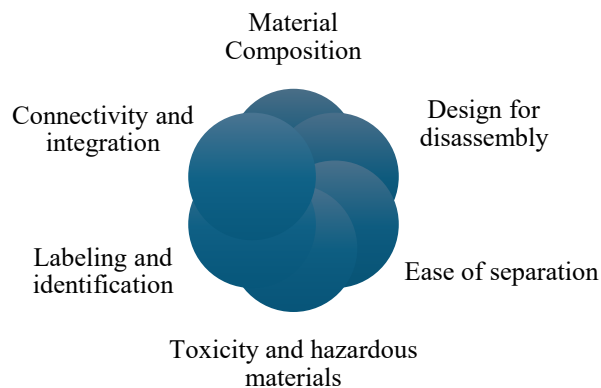

**Fig. 3. Key Components involved in recyclability of WPCBs.**

The concepts of recycling potential, recyclability, and recycling difficulties is very crucial to understand described the overall process of e-waste recycling in terms of economic and speed of recovery. Recycling potential is described as when the amount of revenue exceeds the costs involved in recycling process like collection, transportation, and recycling [29,30]. Recyclability refers to the likelihood that a material can be successfully recycled. It considers the difficulties associated with recycling metals, polymers, and glass using both physical and chemical methods. The actual cost of recycling electronic waste (e-waste) is directly influenced by its recyclability, rather than its mere potential for recycling. In some cases, recyclability can serve as an indicator of economic and technological proficiency. On the other hand, recycling difficulty determines how effectively processes can extract valuable materials from waste. Therefore, the true measure of recycling success for a component lies in its recyclability, not its recycling potential.

*3.2 Measuring the recyclability of WECs present on WPCBs.*

There are multiple challenges involved in the recyclability of electronic components of WPCBs, the first step is to find the material composition of the individual components present on the WPCBs. This information may be provided by the manufacturer of the electronic components but, for confidentiality reasons, this information is not available to the public, except for that required by the EU Restriction of Hazardous Substances in Electrical and Electronic Equipment (RoHS) regulations which limit the use of dangerous substances present in electrical and electronic equipment for the protection of the environment and public health.

Secondly, recyclability state term can only be calculated theoretically to assess the state of recycled materials from WPCBs and identify the level of recycling difficulty. For this we used, innovative method [30] to measure the recyclability of individual components present on the PCBs to facilitate the disassembly and sorting process of WPCBs using Artificial Intelligence. This approach measures recyclability as the theoretical probability of an item being recycled. It specifically focuses on the difficulty of recycling electronics such as metals, plastics, and glass through physical treatment and chemical recovery processes. We applied the same scenarios to the component level of recycling WPCBs.

$$H = -\sum_{i=1}^{n}(P_i \cdot \log_2 P_i) \qquad (1)$$

$$\sum_{i=1}^{n} P_i = 1 \qquad (2)$$

$$D = \sum_{i=1}^{m} D_i = \begin{cases} \sum_{i=1}^{m} P_i, & \text{Physically mixed goods} \\ \sum_{i=1}^{m}[1-(j_i-1)]/N]i, & \text{Chemically Combined goods} \end{cases} \qquad (3)$$

$$R = \frac{100 \cdot D}{N \cdot H} \qquad (4)$$

In above equations, $P_i$ is the probability of the event occurring with material concentration representation $i$, whereas $n$ represents the total number of materials present in a given electronic component; H represents the entropy (measurement of disorder in a system); $D_i$ represents the grade of the material $i$; $m$ the number of materials in a given goods $(m<n)$; $j$ represents the ranking of the materials on the basis of valency, for example copper generally has three valences (e.g. 0, +1, +2); N shows the total number of valences and D is the total grade of the goods with 100 is the coefficient of amplification. R represents the recyclability of the material present in specific electronic components.

*3.3 Case Study*

The initial stage in accurately assessing the recyclability of electronic components involves determining the material composition of WECs and the quality of the material they contain. Many studies have been developed for the material composition of electronic components using conventional recycling methods like hydrometallurgy, pyrometallurgy [4,31]. We utilized data [11] from WECs and developed a circular approach for the automated disassembly of WECs from WPCBs. This study offers detailed evidence of the metal content present in WECs. Each WEC comprises a varying proportion of metals. For example, an Aluminum capacitor contains six distinct types of metals with varying compositions. An increase in the variety of metals in a specific WEC results in decreased recyclability and grade, indicating a lower potential for recycling. We established thresholds for each component's composition and chose five distinct values for grading and recyclability assessments. This provides a comprehensive summary of the recyclability degree of each WECs. The grade of the material content in each component depends on the material type as it is physically or chemically combined structured. Table 1. Shows the classification of each component material as physically or chemically combined or composite.

Table 1. Classification of Component based on the material.

| Component | Type | Reason |
|---|---|---|
| Printed Circuit Board (PCB) | Composite | PCBs consist of layers of various materials, such as conductive (copper) and non-conductive (fiberglass or epoxy resin) layers, that are physically assembled to offer structural support and electrical connections [32]. |
| Aluminum Capacitor | Composite | The composition includes Aluminum foil and a coating of Aluminum oxide, along with a liquid or gel electrolyte. The diverse materials combine physically to enable its function. [33] |
| Tantalum Capacitor | Composite | Composed of Tantalum powder and electrolyte. The powder creates a porous structure with a substantial surface area, covered by an oxide layer serving as a dielectric, and physically mixed with the electrolyte. |
| IC (Integrated Circuit) | Composite | Integrated circuits (ICs) consist of a semiconductor substrate, typically silicon, that is doped with impurities to form p-n junctions, and interconnected by metal channels, frequently made of an alloy. The packaging may contain plastics or ceramics [34]. |
| Diode | Alloy/Composite | Comprised of semiconductor materials like as silicon or germanium that have been doped with impurities, and feature metal contacts that may be alloys. The use of doped semiconductors and metal connections is crucial for its functionality. [34]. |
| Transistor | Alloy/Composite | Transistors, like diodes, are fabricated using semiconductor materials that have been doped with impurities and feature metal (alloy) contacts. These materials can physically influence the flow of electrons [34]. |
| Inductor | Composite | The inductor is made from a copper wire coil wrapped around a core, which can be air, ferrite, or an iron alloy. The wire and core components are physically separate but work together magnetically [32] |
| Resistor | Composite | The material can be created using carbon film, metal film, or metal oxide film applied on an insulating base. Various materials are utilized for the body and the leads, which are physically joined together to form a resistive device [33]. |

## 4    Results and Discussion

*4.1 Recyclability of WECs found on WPCBs.*

The entropy level of each component varies between 0.69 and 2.0. Higher the entropy, lower the potential of recovery of high valued materials. The recycling capability of each WEC varies depending on the material composition of the WECs. Table two shows the entropy, grade of each WECs, and their recyclability. The unit of measurement for entropy is a bit, expressed in certain range and recyclability is /bit. Higher material grade leads to increased high purity of material, its complex composition and conversely. Entropy and recyclability have an inverse connection. High entropy in WECs leads to low recyclability, and vice versa. The recyclability value is determined by the entropy level and the quality of the material.

Table 2. Determination of H, D and R of various electronic components

| Component | H (bit) | D | R(/bit) |
|---|---|---|---|
| Aluminum Capacitor | 0.80 -1.25 | 4.05 | 68 ±14 |
| Tantalum Capacitor | 1.18 – 1.91 | 4.00 | 45 ±11 |
| IC | 1.04 – 1.75 | 3.85 | 33 ±8 |
| Diode | 0.69 – 0.96 | 5.40 | 76 ± 12 |
| Transistor | 0.85 – 1.81 | 3.10 | 58 ± 21 |
| Resistor | 1.51 – 2.00 | 4.15 | 39 ± 6 |
| Inductor | 1.10 – 1.42 | 3.25 | 52 ± 7 |

Fig. 4 shows the recyclability map of each WECs present on WPCBs utilized for automatic disassembly and sorting in the initial stage of recycling. The high Entropy values of each electrical component indicate the complex composition of individual WECs and the challenging recycling process. The recycling rates for WECs range from 10 to 90, which is very low compared to the recyclability rate of 30 for other categories of e-waste. The minimum threshold for recovery is 18 as shown in figure 4, while recyclability value below 30 is considered as difficult to recycle and needs special attention during recycling. If value is from 30-50, it is considered as moderate recycling and from 50 to onward is easy recycling zone. Automatic disassembly and sorting facilitate the efficient recycling of valuable materials from electronic components with minimal loss with respect to recyclability perspective. This also accelerated the recycling process and conserved computational resources on those WECs that are unable to recycle in practical scenarios.

*4.2 Recyclability and Automated Disassembly: A Unified Approach*

We used recyclability criteria to train the advanced deep learning model YOLOv5 on VPCB which is a large dataset of images of electronic components owned by Vega Research Laboratories srl. The VPCBs dataset comprises high-resolution images of WPCBs taken under recycling circumstances. We used Python 3.8 together with an NVIDIA Quadro RTX5000 16GB GPU equipped with 3072 CUDA cores for training and inference. CUDA is a parallel computing platform developed by NVIDIA that leverages GPU architectures to enhance graphics operations and data processing. GPUs are utilized for generating images, sceneries, or rendering 3D visuals, and can accelerate machine learning applications. Table 3 shows the detailed implementation of YOLOv5 model. Fig. 5 and Fig. 6 illustrate the automated identification and localization of every electrical component on WPCBs. The metrics used to assess the detection model is mean average precision. In later phases, the position of each component in a robotic manipulator is utilized to disassemble and categorize the WECs for the recovery of high-end materials. Using recyclability measurements can improve the efficiency of recycling processes by automatically disassembling electronic components on WPCBs. This method enhances resource efficiency by prioritizing high-value WECs, decreases environmental effects by strategic material retrieval, and boosts economic gains by reducing inefficient processing. Furthermore, it is in line with the latest recycling technology to ensure materials are processed efficiently, thereby contributing to sustainable development objectives in e-waste management.

Table 3. Model training and implementation details.

| Parameters | Description |
| --- | --- |
| Framework | YOLOv5 |
| Backbone | CSPDarknet-53 |
| Training Images | 746 |
| Image Size | 1920x1080 |
| Learning Rate | 10-5 |
| Optimizer | Adam |
| Batch Size | 4 |
| Epochs | 150 |

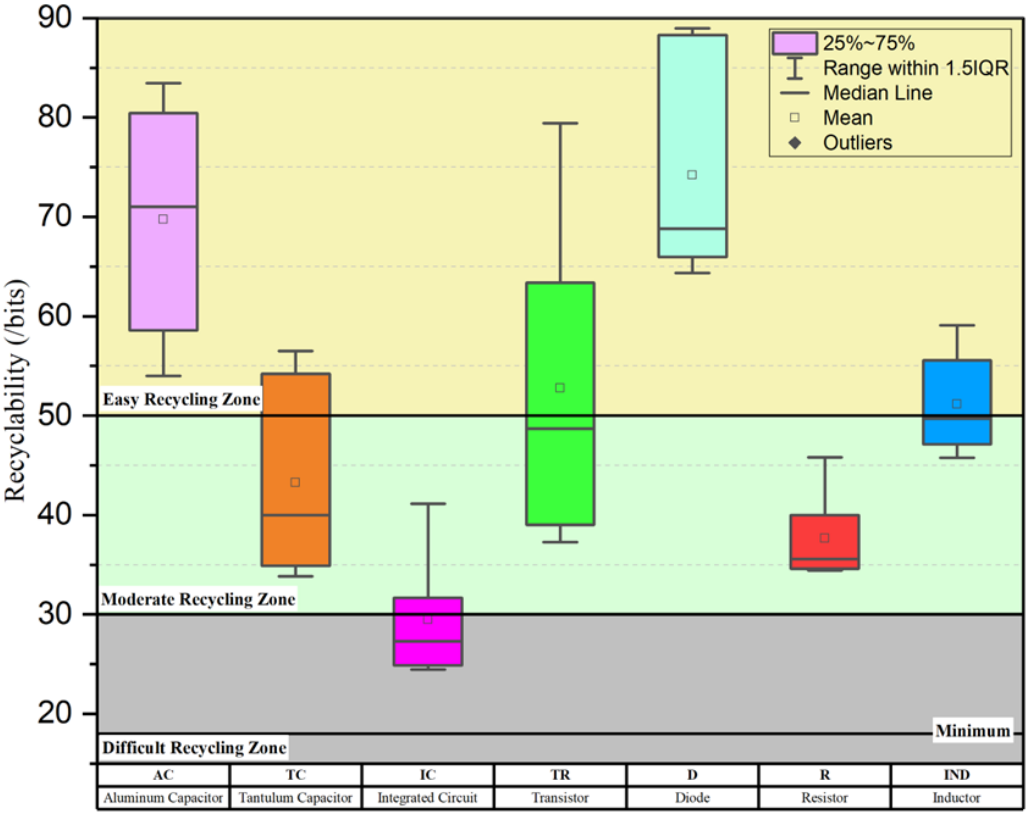

**Fig. 4. Recyclability of map of WPCBs electronic component, each bar shows the recyclability range of each electronic component based on material.**

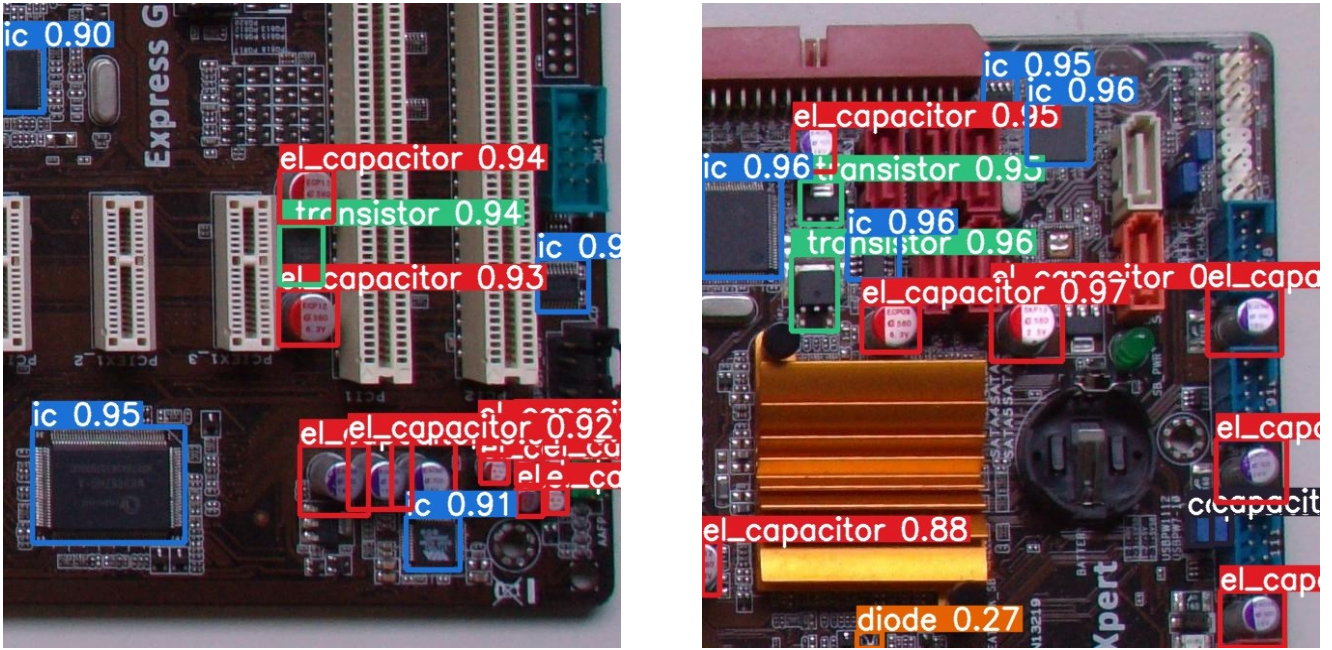

**Fig. 5. Automatic detection and localization of WECs using Artificial Intelligence based Model on V-PCBs dataset.**

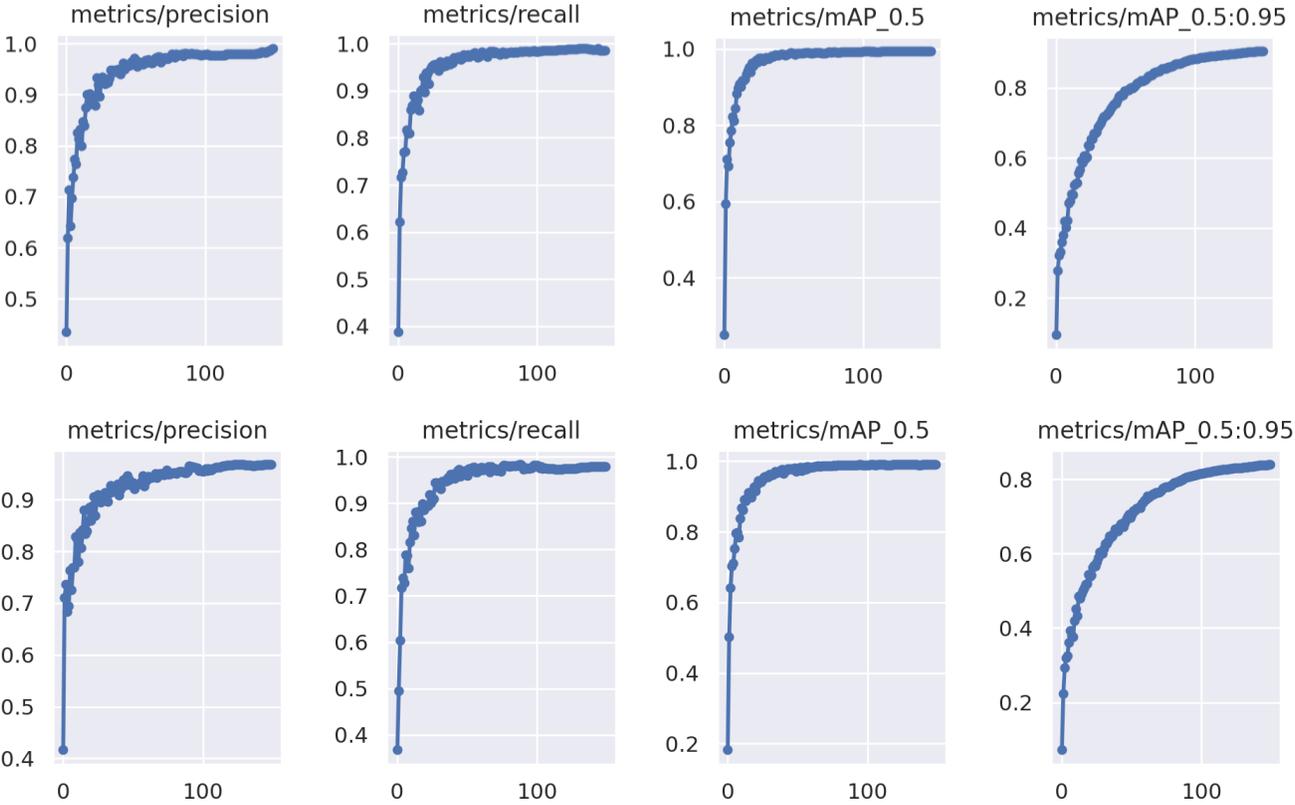

**Fig. 6. Training accuracy of YOLOv5n and YOLOv5x models on V-PCBs dataset.**

## 5    Conclusion

This study utilized a novel approach to determine the recyclability of electronic components present on waste printed circuit boards (WPCBs). We incorporated recyclability criteria at the beginning of the automated disassembly and sorting process to leverage the importance of choosing high-value materials for their economic and environmental impact. The recyclability measurement played a crucial role in training a deep learning object detection model, specifically YOLOv5, with the V-PCB dataset to optimize the recycling process. The model showed cutting-edge performance by detecting components with significant potential for recyclability, crucial for recovering high-value materials. Localization of each individual high recyclable component allows robotic manipulators to efficiently dismantle and sort components, which leads to improve material recovery. Enhancing individual component analysis and including recyclability indicators could significantly enhance the effectiveness of recycling systems, fulfilling a crucial requirement in sustainable waste management.